\documentclass[sigconf]{acmart}

\AtBeginDocument{%
  }

\usepackage{amsmath}
\usepackage{soul}
\DeclareUnicodeCharacter{211D}{\mathbb{R}}
\usepackage{amsthm}
\usepackage{graphicx}
\usepackage{subfig}

\usepackage[ruled,linesnumbered]{algorithm2e}
\usepackage{cleveref}
\crefname{section}{§}{§§}
\theoremstyle{plain}
\newtheorem{assumption}{Assumption}

\newtheorem{theorem}{Theorem}

\newtheorem{lemma}{Lemma}

\copyrightyear{2024}
\acmYear{2024}
\setcopyright{acmlicensed}\acmConference[KDD '24]{Proceedings of the 30th ACM SIGKDD Conference on Knowledge Discovery and Data Mining}{August 25--29, 2024}{Barcelona, Spain}
\acmBooktitle{Proceedings of the 30th ACM SIGKDD Conference on Knowledge Discovery and Data Mining (KDD '24), August 25--29, 2024, Barcelona, Spain}
\acmDOI{10.1145/3637528.3671892}
\acmISBN{979-8-4007-0490-1/24/08}
\begin{document}
\acmSubmissionID{rtp1332}
\title{Conversational Dueling Bandits in Generalized Linear Models}

\author{Shuhua Yang}
\affiliation{%
  \institution{University of Science and Technology of China}
  \city{Hefei}
 \country{China}}
\email{shuashua0608@mail.ustc.edu.com}

\author{Hui Yuan}
\affiliation{%
  \institution{Princeton University}
  \city{Princeton}
  \state{NJ}
  \country{USA}}
\email{huiyuan@princeton.edu}

\author{Xiaoying Zhang}
\affiliation{%
  \institution{ByteDance}
  \city{Beijing}
  \country{China}}
\email{zhangxiaoying.xy@bytedance.com}

\author{Mengdi Wang}
\affiliation{%
  \institution{Princeton University}
   \city{Princeton}
   \state{NJ}
  \country{USA}}
\email{mengdiw@princeton.edu}

\author{Hong Zhang}
\affiliation{%
 \institution{University of Science and Technology of China}
 \city{Hefei}
 \country{China}}
\email{zhangh@ustc.edu.cn}

\author{Huazheng Wang}
\affiliation{%
  \institution{Oregon State University}
  \city{Corvallis}
  \state{OR}
  \country{USA}}
\email{huazheng.wang@oregonstate.edu}
\renewcommand{\shortauthors}{Shuhua Yang et al.}
\begin{abstract}
 
Conversational recommendation systems elicit user preferences by interacting with users to obtain their feedback on recommended commodities. Such systems utilize a multi-armed bandit framework to learn user preferences in an online manner and have received great success in recent years. However, existing conversational bandit methods have several limitations. First, they only enable users to provide explicit binary feedback on the recommended items or categories, leading to ambiguity in interpretation. In practice, users are usually faced with more than one choice. Relative feedback, known for its informativeness, has gained increasing popularity in recommendation system design. 
Moreover, current contextual bandit methods mainly work under linear reward assumptions, ignoring practical non-linear reward structures in generalized linear models. 
Therefore, in this paper, we introduce relative feedback-based conversations into conversational recommendation systems through the integration of dueling bandits in generalized linear models (GLM) and propose a novel conversational dueling bandit algorithm called ConDuel. Theoretical analyses of regret upper bounds and empirical validations on synthetic and real-world data underscore ConDuel's efficacy. We also demonstrate the potential to extend our algorithm to multinomial logit bandits with theoretical and experimental guarantees, which further proves the applicability of the proposed framework. 
\end{abstract}



\maketitle

\section{Introduction}

Contextual bandit is an essential tool in recommendation systems to enhance the performance of the system while making a trade-off between exploitation and exploration \cite{li2010contextual,abbasi2011improved}. In recommendation scenarios, each item is considered as an arm with its contextual vector summarizing the information of both the arm and the user. At each round, the recommendation system sequentially suggests items to the user and collects feedback (e.g., click) on the selected item.
The agent's goal in the system is to develop an item recommendation (arm selection) strategy that maximizes the cumulative reward from the user by leveraging information about the user and items, as well as the user's previous interaction records. 

In many scenarios, it is challenging to effectively utilize user feedback and recommend optimally for cold-start users due to limited historical data, with insufficient data to learn users' preferences reliably. To accelerate the learning process of user preferences and offer optimal recommendations, conversational recommendation systems (CRSs) have been proposed in \cite{christakopoulou2016towards}, \cite{zhang2020conversational} and \cite{xie2021comparison}.
In CRSs, the system not only gathers responses on recommended items but also sparks conversations by asking users about relevant "key-terms," such as categories or entities associated with news articles in news recommendation systems. According to \cite{zhang2020conversational}, these interactions accelerate CRS learning by leveraging key-terms linked to numerous items, offering valuable insights into user preferences. 

\begin{figure}
\centering
\includegraphics[width=0.95\linewidth,page=3]{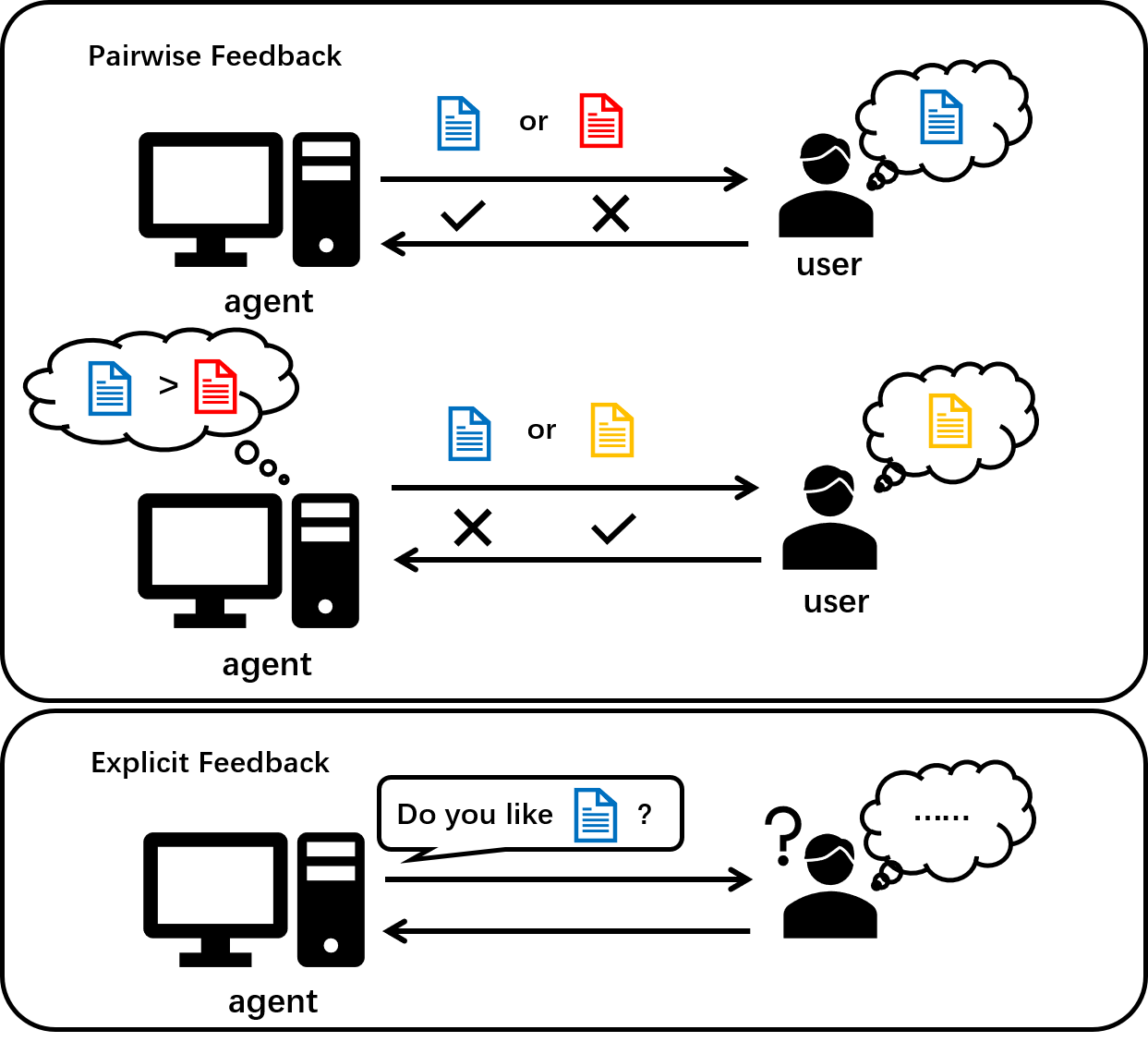}
\caption{An illustrative example of a conversational system with pairwise feedback compared with a system that only allows explicit feedback.}
\Description{Illustrate pairwise feedback mechanism is better and more practical than explicit feedback mechanism.}
\label{fig:1}
\end{figure}

Despite previous successes in CRSs, current conversation mechanisms, particularly conversational contextual bandit approaches, often fall short. First and foremost, current conversational contextual bandit approaches concentrate solely on explicit user feedback for specific items/categories, which can be ambiguous and fail to effectively capture user preferences. In contrast, relative feedback has been proven to be informative and is commonly observed in various settings including recent applications in Reinforcement Learning with Human Feedback (RLHF) \cite{ouyang2022training, zhu2023principled, sekhari2023contextual, ji2023provable}.
To better illustrate the difference between CRS with different feedback mechanisms, we give a simple yet illustrative example in Fig~\ref{fig:1}. For previous CRSs that only allow explicit feedback, they would inquire about the user's preference on a particular commodity category through queries such as:  "Are you interested in digital products?", making it difficult for the user to respond with a simple "Yes" or "No", especially when the user is unsure about the specific type of digital products being asked about. The user may be interested in Ebooks but dislikes video games, thus providing ambiguous feedback to the agent. Alternatively, for agents that allow relative feedback and ask questions such as "Do you prefer digital products or video games?", the user can provide more decisive feedback, prompting the system to understand the user's preference more efficiently. 

Building on these observations, our paper proposes to build a CRS that guarantees relative feedback within the dueling bandit framework \cite{yue2012k, yue2011beat, zoghi2014relative}. Inspired by \cite{dudik2015contextual}, which proposed the first contextualized extension of dueling bandits, we also incorporate the contextual information into this framework.
Furthermore, although most CRSs have adopted the linear bandit framework,  the linear reward assumptions may not align with practical scenarios. To address this limitation, we relax the linear reward assumption and design a more practical dueling bandit approach by employing a generalized linear model (GLM). This approach, as demonstrated, yields significant improvements over the linear models commonly utilized in current CRSs \cite{li2012unbiased}.
Designing a CRS under a dueling bandits framework in GLM imposes challenges, including (1) determining key-term pairs to query and (2) selecting arm pairs based on interactions from both the key-term and arm modules. To address these, we propose the Conversational Dueling Bandit algorithm (ConDuel). Our approach involves conversations on "exploratory" key-term pairs and item pair selection based on the uncertainty principle.
Leveraging feedback from both modules, our ConDuel algorithm extends CRS to a pairwise dueling bandit model. Through experiments on synthetic and real-world datasets, we demonstrate the competitiveness of our algorithm over baselines. We also demonstrate the potential to extend the pairwise feedback model into a multi-choice model, proposing a ConMNL algorithm that can tackle the multinomial logit bandit problem. In summary, the contribution of our work is three-fold:
\begin{itemize}
    \item We propose a new framework for conversation recommender systems (CRS) that can efficiently utilize relative feedback upon each query. We specifically design the dueling bandit algorithm ConDuel to achieve the objective. 
    To the best of our knowledge, this is the first work that enables informative pairwise preference-based questions on both key terms and items in CRS. 
    \item Our ConDuel algorithm applies a generalized linear model and we provide a sublinear regret upper bound as theoretical support. Besides pairwise comparison, we also extend ConDuel to multiple comparisons under the choice model with the proposed ConMNL algorithm.
    \item Extensive experiments on a synthetic dataset and two real-world datasets verify the efficiency of the proposed ConDuel algorithm and ConMNL algorithm.
    \end{itemize}

\section{Related work}
Our work builds on several research areas, and we review some recent work in the most related areas.
\noindent\paragraph{Conversational Bandits.}  Contextual bandit algorithms aim to optimize the expected cumulative rewards, in the long run \cite{li2010contextual,abbasi2011improved}. Traditional linear bandits require extensive exploration to learn user preferences in recommender systems. \cite{christakopoulou2016towards} first proposed multi-armed bandit models in Conversational recommender systems to acquire users' feedback on each item. Afterward, \cite{zhang2020conversational} systematically studied conversational contextual bandit and proposed a ConUCB algorithm to accelerate online recommendations,  which allows the agent to obtain user feedback on key-terms related to items and leverage this information to accelerate the system. Building on this framework, some follow-up works have extended conversational contextual bandits in various settings, such as using clustering techniques to create self-generated key-terms \cite{wu2021clustering}, obtaining relative feedback from key-terms \cite{xie2021comparison}, leveraging knowledge graphs to study the underlying relations between key-terms \cite{zhao2022knowledge}, and incorporating both information from arm-level and key-term-level to construct a holistic model~\cite{wang2023efficient}.
Although recent work in \cite{xie2021comparison} has incorporated relative feedback in the key-term module, the algorithm proposed in their work is a simple empirical extension of ConUCB in \cite{zhang2020conversational}, in which the system utilizes a pseudo update over key-terms without theoretical guarantees.
 

\noindent\paragraph{Utility-based Dueling Bandits.} In utility-based dueling bandits, the absolute preference for each arm can be reflected by a real-valued utility degree \cite{bengs2021preference}. When applied to dueling bandits, this setting is also known as \textit{utility-based dueling bandits}, where a latent utility function $u: \mathcal{A}\to \mathbb{R}$ exists, with $u(a_i)$ representing the utility of an arm $a_i\in\mathcal{A}$. The probability of arm $a_i$ winning over $a_j$ can be determined by the difference of their utilities using a link function $\mu: \mathbb{R}\to [0,1]$, and can be written as $P(a_i>a_j) = \mu(u(a_i) - u(a_j))$. In contextual bandit, the utility of an arm is assumed to be linear based on an unknown preference vector, which has been studied in \cite{saha2021optimal,saha2022efficient,bengs2022stochastic}. Utility-based dueling bandits can also be extended to the preference-based reinforcement learning framework in \cite{xu2020preference} and \cite{pacchiano2021dueling}.

\section{Problem Formulation}\label{sec:prob}

In this section, we introduce the general framework of conversational dueling bandits with a generalized linear model (GLM). Suppose there are $N$ arms denoted by $\mathcal{A}$ and $K$ key-terms denoted by $\mathcal{K}$. At each round $t=1,...,T$, the agent is given a subset of arms $\mathcal{A}_t\subset\mathcal{A}$, where each arm $a\in\mathcal{A}_t$ is associated with a contextual vector $x_{a,t}\in\mathbb{R}^d$. Without loss of generality, we assume that the feature vectors are normalized, i.e., $\|x_{a,t}\|_2 = 1$. We also assume that the unknown user preference vector $\theta_*\in\mathbb{R}^d$ satisfies the inequality $\|\theta_*\|_2\leq 1$. The relationship between the arms and the key-terms can be characterized by a weighted bipartite graph $(\mathcal{A},\mathcal{K},\mathcal{W})$, whose nodes are divided into two sets $\mathcal{A}$ and $\mathcal{K}$, and weighted edges are represented by the matrix $\mathcal{W} \triangleq [w_{a,k}]$ with $w_{a,k}$ representing the relationship between arm $a$ and key-term $k$. Without loss of generality, we assume $\sum_k w_{a,k}= 1$.  

\paragraph{Generalized Linear Dueling Feedback.} 
We assume the latent utility function $u: \mathcal{A}\to \mathbb{R}$ is linear and $u(a, t) = x_{a,t}^T\theta_*$ represents the utility of the arm $a\in\mathcal{A}_t$ at round $t$, with $\theta_*\in\mathbb{R}^d$ being the unknown preference vector. We also define a link function $\mu: \mathbb{R}\to [0,1]$, so that at round $t$, the probability of arm $a_i\in\mathcal{A}_t$ winning over $a_j\in\mathcal{A_t}$ satisfies the following equation:
\begin{align}\label{eq:1}
    P(a_i>a_j) = \mu(u(a_i, t) - u(a_j,t)) = \mu(x_{i,t}^T\theta_* - x_{j,t}^T\theta_*).
\end{align}
The link function $\mu$ satisfies the following properties~\cite{yue2012k}:
\begin{itemize}
    \item $\mu$ is monotonically increasing, so that an arm with a higher utility than another arm will have a higher probability to be chosen than the latter.
    \item $\mu(0) = 1/2$, indicating that two arms having the same utility have also the same probability of being chosen.
    \item $\mu(-\infty)=0$, $\mu(\infty)= 1$.
\end{itemize}
It is easy to verify the two most common link functions: logistic function $\mu(x) = 1/(1+\exp(-x))$ and linear function $\mu(x) = \max\{0, \min\{1,0.5*(1+x)\}\}$ both satisfy the properties.
Following \cite{filippi2010parametric}, we assume that $\mu$ satisfies the following assumptions.

\begin{assumption} \label{ass1}
$\kappa_1 = \inf_{\{\|x\|_2\leq 2, \theta\in\Theta\}}\mu^{\prime}(x^T\theta)>0$, where $\Theta$ is a closed subset of the space $\mathbb{R}^d$ containing $\theta_*$.
\end{assumption}

\begin{assumption} \label{ass2}
$\mu$ is twice differentiable, and its first and second-order derivatives are upper-bounded by constant $L_{\mu}$ and $M_{\mu}$ respectively. When $\mu$ is the sigmoid function, $L_{\mu}$ and $M_{\mu}$ can be $1/4$.
\end{assumption}


Similar to Eq~\ref{eq:1}, we define the probability of key-term $k_i$ beating $k_j$ at round $t$ as
\begin{align}
    P(k_i>k_j) = \mu(\Tilde{u}(k_i,t) - \Tilde{u}(k_j,t)),\notag
\end{align}
where $\Tilde{u}(k,t):= \frac{\sum _{a\in\mathcal{A}}w_{a,k}u_(a,t)}{\sum_{a\in\mathcal{A}} w_{a,k}}$, indicating that the utility of $k$ is determined by averaging over that of its related arms.
The relative feedback of two key-terms in a duel is also determined by their utilities. Equivalently, we rewrite the inequality as 
\begin{equation}\label{eq:2}
    P(k_i>k_j) = \mu(\Tilde{x}_{k_i, t}^T\theta_*-\Tilde{x}_{k_j, t}^T\theta_*),
\end{equation}
with $\Tilde{x}_{k,t} =\frac{\sum _{a\in\mathcal{A}}w_{a,k}x_{a,t}}{\sum_{a\in\mathcal{A}} w_{a,k}}$ representing the feature vector of key-term $k$ at round $t$.

At round $t$, given the candidate arm set $\mathcal{A}_t$, we present to the user a pair of arms $(a_t, a_t^{\prime}) \in \mathcal{A}_t\times \mathcal{A}_t$ and ask for his/her relative preference. At round $t$, user's preference is encoded by a binary random variable $o_t = \textbf{1}(a_t>a_t^{\prime})$. Denote by $d_t: = x_{a_t}-x_{a_t^{\prime}}$, $o_t$ follows the Bernoulli distribution $Ber(\mu(d_t^T\theta_*))$, and the arm-level feedback model can be written as
\begin{equation}\label{eq:3}
    o_t = \mu(d_t^T\theta_*)+\epsilon_t,
\end{equation}
where $\epsilon_t$ is a zero-mean noise defined as
$$
\epsilon_t =\left\{
\begin{aligned}
1-\mu(d_t^T\theta_*) & , & \text{with probability } \mu(d_t^T\theta_*), \\
-\mu(d_t^T\theta_*) & , & \text{with probability } 1-\mu(d_t^T\theta_*).
\end{aligned}
\right.
$$
It is easy to verify that $\epsilon_t$'s are $R$-sub-Gaussian with $R \leq 1/2$.

\paragraph{Conversation on Key-Terms and Frequency} 
CRS obtains additional user feedback through additional key-term conversations. Similarly, the key-term level feedback on comparing $(k_t,k_t^{\prime})$ satisfies $\tilde{o}_t \sim Ber(\Tilde{d}_t^T\theta_*)$, where $\Tilde{d}_t = \Tilde{x}_{k_t,t}-\Tilde{x}_{k_t^{\prime},t}$. The key-term level model is presented as
\begin{align}\label{eq:4}
\tilde{o}_t =\mu(\Tilde{d}_t^T\theta_*)+\Tilde{\epsilon}_t.
\end{align}
$b(t)$ is introduced to model the frequency of conversations, and we consider the following function: 
$$q(t) = \left\{
\begin{aligned}
    1, \quad & b(t)-b(t-1)>0,\\
    0, \quad & \text{otherwise}.
\end{aligned}
\right.$$
The agent conducts $\lfloor b(t)-b(t-1)\rfloor$ conversations with the user at round $t$ when $q(t)=1$ and refrains conversations when $q(t)=0$. We also assume the key-term-level conversations are less frequent than arm-level interactions, ensuring $b(t)\leq t$ for any $t$ to prioritize users' experience. 

\paragraph{Cumulative Regret.}
At round $t$, denote the best arm as $a_t^{*}$, with $a_t^{*} = \arg\max_{a\in\mathcal{A}_t}u(a,t) = \arg\max_{a\in\mathcal{A}_t} x_{a,t}^T\theta_{*}$, and the the chosen arms pair as $({a_t}, {a_t^{'}})$. Following \cite{bengs2022stochastic}, the instantaneous dueling bandit regret is defined as
\begin{align}
    r_t & = u(a_t^{*},t) - \frac{1}{2}(u(a_t, t)+u(a_t^{'},t))\notag\\
    & = x_{a_t^{*}, t}^T\theta_*-\frac{1}{2}(x_{a_t}^T\theta_* + x_{a_t^{'}}^T\theta_*)\notag
\end{align}
The cumulative dueling bandit regret is defined as
\begin{align}
\begin{split}\label{eq:5}
    R(T)& = \sum_{t=1}^T \left(u(a_t^{*},t) - \frac{1}{2}(u(a_t, t)+u(a_t^{'},t))\right)\\
    & = \sum_{t=1}^T\left(x_{a_t^{*}, t}^T\theta_*-\frac{1}{2}(x_{a_t}^T\theta_*+ x_{a_t^{'}}^T\theta_*)\right).    
\end{split}
\end{align}

\section{Conversational Dueling Bandits with GLM and Regret Analysis} 
We propose the Conversational Dueling Bandits algorithm (ConDuel) (Alg. \ref{alg}) to address the three challenges in the conversational dueling bandit setting: 1) how to exploit the historical feedback from both arms and key-terms. 2) how to select key-term pairs for conversation to explore better; 3) how to select arm pairs to mostly minimize the cumulative regret. In \cref{sec:pipe}, we introduce the full algorithm and discuss in detail the highlights of ConDuel addressing the three challenges above. Meanwhile, theoretical analysis for ConDuel is provided and a regret upper bounds is showcased in section \cref{sec:rgt}.

\subsection{ConDuel Algorithm}\label{sec:pipe}
 
We now detail the proposed ConDuel algorithm as follows: 
\begin{itemize}
    \item \textbf{Key-term selection module.} If the agent conducts a conversation with the user based on the previous interactions, it will select a pair of 'explorative' key-terms (which will be discussed in detail later) to query. 
    Then the collected feedback from the key-term level will be passed to the full model. We further assume that $b(t) = b\cdot t$ in the following analysis for simplification, where $b\in (0,1).$ 
    \item \textbf{Arm-selection module.} Based on the previous interaction history, the parameter $\theta_t$ is calculated and maintains an optimistic estimate on the dueling feedback $o_t$ based on the UCB principle. 
    The agent subsequently constructs a subset $C_t$, containing promising arms that are likely to be optimal;
    Then the system selects a pair of arms from $C_t$ that are most uncertain to explore the users' preference thoroughly and updates the parameters based on the user's choice. 
\end{itemize}
Based on the arm-level feedback and conversational feedback received in the previous rounds, the two modules interact with the users. The agent utilizes feedback from both modules for recommendations. 
The main difference between our proposed algorithm and ConUCB lies in the usage of relative feedback from both the key-term level and arm level, guaranteed by the introduction of a maximum likelihood estimator (MLE).
To the best of our knowledge, this is the first non-trivial extension of ConUCB in the generalized linear model with dueling key-terms module. We present the ConDuel algorithm in \textbf{Algorithm} \ref{alg}. The main body of ConDuel contains a key-term selection module (lines 3-11) and an arm selection module (lines 14-16). Note that at each iteration $t$, we try to maintain a tight estimate $\theta_t$ of the true parameter $\theta_{*}$ utilizing feedback from both key-term level interactions and arm-level interactions in our generalized linear model.


\begin{algorithm}[ht]
\SetAlgoLined
\KwIn{ $(\mathcal{A},\mathcal{K},\mathcal{W}),b(t), \lambda, \kappa_1$;}
\textbf{Initialization}: $M_0 = \frac{\lambda}{\kappa_1}I$ \; 
\For{$t$ = 1,...,T}{\eIf{$b(t)-b(t-1)>0$}{
$q_t=b(t)-b(t-1)$\;
\While{$q_t>0$}{
select a pair of key-terms $(k_t,k_t^\prime)$ independently from barycentric spanner $\mathcal{B}$\;
Receive relative feedback $\Tilde{o}_t = \textbf{1}(k_t>k_t^\prime)$, $\Tilde{d_t} = \Tilde{x}_{k_t}-\Tilde{x}_{k_t^\prime}$\;
Update ${M}_t={M}_{t-1}+\Tilde{d_t}\Tilde{d_t}^T$\;
$q_t= q_t - 1$;}}
{$M_t = M_{t-1}$}
$\theta_t$ is estimated based on Eq.\eqref{eq:7} or Eq. \eqref{eq:8}, $\theta_t^{(1)}$ is computed according to Eq. \eqref{eq:10}\;
Construct $C_t= \{a\in \mathcal{A}_t|(x_{a,t}-x_{a^\prime, t})^T\theta_t^{(1)}+\alpha_t \|x_{a, t}- x_{a^\prime,t}\|_{M_t^{-1}}>0, \forall a^\prime\in \mathcal{A}_t\}$\;
Select the arm pair $a_t$ and $a_t^{\prime}$ from $\mathcal{C}_t$ satisfying:
$(a_t, a_t^{\prime}) = \arg\max_{a,a^{\prime}\in\mathcal{C}_t}\{\|x_{a,t}-x_{a^{\prime},t}\|_{M_t^{-1}}\}$\;
Receive feedback $o_t = \textbf{1}(a_t>a_t^{\prime})$, $d_t = x_{a_t, t} - x_{a_t^{\prime}, t}$\;
Update $M_t = M_{t-1} + d_t d_t^T$}
\caption{The ConDuel Algorithm}
\label{alg}
\end{algorithm}
\subsubsection{Parameter Estimation}

Previous CRSs estimate $\theta_*$ separately from arm-level and key-term-level~\cite{zhang2020conversational, xie2021comparison, wu2021clustering}, which may cause waste of observations. In our model, we fully utilize information from both arm-level feedback and key-term-level feedback to obtain the MLE of $\theta_*$ by solving one optimization problem that maximizes the log-likelihood function. Based on our model \eqref{eq:3} and \eqref{eq:4}, the regularized MLE of $\theta_*$ in our model is given by
\begin{align}
\begin{split}\label{eq:6}
\theta_t\in&\arg\max_{\theta\in\Theta}\{\sum_{s=1}^{t-1}(o_s d_s^T\theta - m(d_s^T\theta))\\
+&\sum_{s=1}^t\sum_{\Tilde{d}_s\in \tilde{\mathcal{D}}_s}(\Tilde{o}_{s} \Tilde{d}_{s}^T\theta - m(\Tilde{d}_{s}^T\theta))-\frac{\lambda}{2}\|\theta\|_2^2\}.
\end{split}
\end{align}
We denote $d_t = x_{a_t,t} - x_{a_t^{\prime},t}$ as the difference contextual vector for the chosen arm pair $({a_t}, {a_t^{\prime}})$, and $\tilde{d}_t = \Tilde{x}_{k_t,t}-\Tilde{x}_{k_t^{\prime},t}$ as the difference contextual vector for the key-term pair $({k_t},{k_t^{\prime}})$ being queried at round $t$. We also define the $\mathcal{D}_t = \{d_t|d_t = x_{a_t,t}-x_{a_t^{\prime},t}, \forall a_t, a_t^{\prime}\in\mathcal{A}_t\}$ and $\mathcal{\Tilde{D}}_t = \{\Tilde{d}_t| \Tilde{d}_t = \Tilde{x}_{k_t,t} - \Tilde{x}_{k_t^{\prime},t}, \forall k_t, k_t^{\prime} \in {\mathcal{K}}_t\}$ at round $t$.  Since the log-likelihood function is strictly concave in $\theta$, the regularized MLE $\theta_t$ in our model is the unique solution of the following score equation upon differentiating:
\begin{equation}\label{eq:7}
\sum_{s=1}^{t-1}(o_s - \mu(d_s^T\theta))d_s+\sum_{s=1}^t\sum_{\Tilde{d}_s\in\tilde{\mathcal{D}}_s}(\Tilde{o}_{s} - \mu(\Tilde{d}_{s}^T\theta))\Tilde{d}_{s}-\lambda\theta=0.
\end{equation}

According to Eq.\eqref{eq:6} and Eq. \eqref{eq:7}, we can define the invertible function $g_t(\theta)$ and the design matrix $M_t$ as
\begin{align}\label{eq:8}
    &g_t(\theta) =  \sum_{s=1}^{t-1} \mu(d_s^T\theta)d_s+\sum_{s=1}^t\sum_{\Tilde{d}_s\in\tilde{\mathcal{D}}_s}\mu(\Tilde{d}_{s}^T\theta)\Tilde{d}_{s}+\lambda \theta,\notag\\
    & M_t = \sum_{s=1}^{t-1}d_s d_s^T+\sum_{s=1}^t\sum_{\Tilde{d}_s\in\tilde{\mathcal{D}}_s}\Tilde{d}_s\Tilde{d}_s^T+\lambda/\kappa_1 I. 
\end{align}

In case the MLE ${\theta}_t$ is outside of the parameter space $\Theta$, we need to add a projection step to obtain ${\theta}_t^{(1)}$ by the techniques in \cite{filippi2010parametric}:
\begin{equation}\label{eq:9}
    \theta_t^{(1)} = \arg\min_{\theta\in\Theta}\|g_t(\theta) - g_t(\theta_t)\|_{M_t^{-1}}.
\end{equation}
Note that when $\theta_t\in \Theta$, we set $\theta_t^{(1)}=\theta_t$ to save the computation. 

\subsubsection{Arm Selection Module} 
\label{sec:arm}
In this part, we will give a detailed description of arm pair selection. At round $t$, for any $a_t, a_t^{\prime}\in\mathcal{A}_t$, our algorithm calculates the UCB estimate on the pairwise feedback:
$$s(a_t, a_t^{\prime}) = (x_{a_t,t} -x_{a_t^{\prime},t})^T\theta_t^{(1)} + \alpha_t\|x_{a_t,t} -x_{a_t^{\prime},t}\|_{M_t^{-1}},$$ then the agent constructs a subset $C_t$ which contains all the promising arms that are superior to the rest of the arms in terms of UCB estimate. The selected pair of arms $(a_t, a_t^{\prime})$ satisfies: $(a_t,a_t^{\prime}) =\arg\max_{a,a^{\prime}\in\mathcal{C}_t}\|x_{a,t}-x_{a^{\prime},t}\|_{M_t^{-1}}$. In this way, we can eliminate arms that are unlikely to be optimal in the first step, and then select the maximum informative arm pair. Notice that this arm selection strategy strictly follows \cite{saha2021optimal}, and when $|\mathcal{C}_t|$ is large, this step can bring a lot of computation. Therefore in our experiment, the first arm $a_t$ is randomly sampled from $C_t$, and the second arm $a_t^{\prime}$ is defined as $a_t^{\prime} = \arg\max\|x_{a_t^{\prime},t}-x_{a_t,t}\|_{M_t^{-1}}$, which can be seen as the most uncertain arm to compare with the first arm. 

\begin{lemma}\label{lemma:1}
Assume $\epsilon_t$ and $\Tilde{\epsilon}_t$ defined in Eq. \eqref{eq:3} and \eqref{eq:4} are conditional $R$-sub-Gaussian, $d_t$ is denoted as the difference contextual vectors for the selected arm pair $(x_t, x_t^{\prime})$.
Then for any $d_{t}\in\mathcal{D}_t$, with probability at least $(1-\delta)$, we have the following inequality:
\begin{equation}\label{eq:10}
    |d_t^T\theta_t^{(1)}-d_t^T\theta_*| \leq \alpha_t\|d_t\|_{M_t^{-1}},
\end{equation}
where $\alpha_t = \frac{2}{\kappa_1}(R\sqrt{d\log((1+\frac{4\kappa_1(t+b(t))}{d\lambda})/\sigma)}+\sqrt{\lambda\kappa_1}\|\theta_*\|_2).$
\end{lemma}

\subsubsection{Key-term Selection Module}
\label{sec:keyterm}
In this section, we describe how the algorithm selects key-term pairs. We hope that the key-term selection module is explorative, that is, to ask questions on key-terms that accelerate the learning of user preferences. Especially, We propose a new strategy for selecting key-term pairs from the barycentric spanner $\mathcal{B}$ from the key-term set $\mathcal{K}$, which aims at exploring key-term information efficiently.

\noindent\textbf{Definition of Barycentric Spanner}. According to \cite{awerbuch2004adaptive}, the subset $\mathcal{B} = \{k_1,...,k_d\}$ is a barycentric spanner for key-term set $\mathcal{K}$, if every $k\in\mathcal{K}$ can be expressed as a linear combination of elements of $\mathcal{B}$ using coefficients in $[-1,1]$, i.e., $\tilde{x}_k = \sum_{i=1}^d c_i\tilde{x}_{k_i}$ ($c_i\in [-1,1]$).

We assume $\mathcal{K}_t$ spans $\mathbb{R}^d$ at each round, thus the constructed barycentric spanner $\mathcal{B}_t$ forms the basis for $\mathbb{R}^d$. In $t$-th round conversation, we sample a pair of key-terms $k_1, k_2\sim \mathcal{B}_t$ independently from the barycentric spanner to obtain relative feedback. 
This is efficient in computation because reducing the number of key-terms can bring a lot of convenience. Based on the definition of barycentric spanner, all information of $\mathcal{K}_t$ can be seen contained in the barycentric spanner, therefore exploring $\mathcal{B}_t$ is sufficient in collecting user feedback.
Furthermore, this strategy can also guarantee some good properties in our algorithm, ensuring a high probability lower bound of $\lambda_{\min}(M_t)$ as follows:

\begin{lemma}\label{lemma:2}
Let $\Sigma = E_{x,y\sim\mathcal{B}}[(x-y)( x-y)^{T}]$, and $\lambda_B = \lambda_{\min}(\Sigma)$. 
As the conversation frequency $b(t)\leq t$, we assume that $b(t) = b\cdot t$ for some $b\in(0,1)$. Then when $t\geq \frac{4(C_1\sqrt{d}+C_2\sqrt{\log(1/\delta))^2}}{b\lambda_B^2}\triangleq t_0$, with probability at least $(1-\delta)$, we have
\begin{equation}\label{eq:12}
    \lambda_{\min}(M_t)\geq \frac{\lambda_B b t}{2}+\frac{\lambda}{\kappa_1},
\end{equation}
with $C_1$ and $C_2$ being constants.
\end{lemma}


\begin{lemma}\label{lemma:3}
With key-term pair independently sampled from barycentric spanner $\mathcal{B}_t$ at each round, and $t_0$ is defined in Lemma~\ref{lemma:2}, then $\forall t>t_0$, we have the following inequality:
$$\sum_{s = t_0+1}^t\|d_s\|_{M_s^{-1}}\leq 8(\sqrt{\frac{t}{2b\lambda_B}}-\sqrt{\frac{t_0}{2b\lambda_B}})\leq 8\sqrt{\frac{t}{2b\lambda_B}}.$$
\end{lemma}
Note that the above lemma uses a different technique to prove the upper bound for $\sum_{t}\|d_t\|_{M_t^{-1}}$, 
and as the conversation frequency $b$ increases, the regret upper bound decreases accordingly. This tendency corresponds to our understanding of the conversation system: the more questions the agent asks, the more feedback it can leverage, thus the more accurately it can learn the user preferences.

\noindent\textbf{Remark.} Notice that the key-term selection module is regret-free,  we don't need to consider explore-exploit trade-offs and apply the UCB principle here. Instead, utilizing an explorative strategy such as choosing key-terms from the barycentric spanner can improve the performance of the algorithm as well as save computation. In the future, it would be interesting to investigate other explorative strategies from best arm identification literature, such as works in \cite{audibert2010best} and \cite{kveton2020randomized}.

\subsection{Regret Upper Bound}
\label{sec:rgt}

We give the upper bound of the cumulative dueling regret $R(T)$ for our algorithm as follows, where we assume $b(t) = b\cdot t$ for some $b\in (0,1)$ from Lemma~\ref{lemma:2}.

\begin{theorem} \label{thm:regret}
With probability (1-$\delta$), our algorithm has the following regret upper bound:
\begin{align}\label{eq:13}
    R(T)&\leq t_0+\frac{32}{\kappa_1}(R\sqrt{d\log((1+\frac{4\kappa_1(T+ b T)}{d\lambda})/\sigma)}+\notag\\
    &+\sqrt{\lambda\kappa_1}\|\theta_{*}\|_2)\sqrt{\frac{T}{2b\lambda_B}} = \mathcal{O}(\frac{1}{\kappa_1}\sqrt{d T\log(T)})
\end{align}
\end{theorem}

It can be seen that the upper bound of $R(T)$ decreases as $b$ increases. As far as we know, this is the first work in a conversational recommender system that directly shows the impact of conversations and proves an explainable regret upper bound concerning conversation frequency. While there’s no direct lower bound of the conversational dueling bandit problem, \cite{saha2021optimal} contained the regret lower bounds of contextual dueling bandits that can also match with ours, i.e., $\Omega(\sqrt{dT})$. 

\begin{figure*}
    \centering   
 \includegraphics[width=\textwidth]{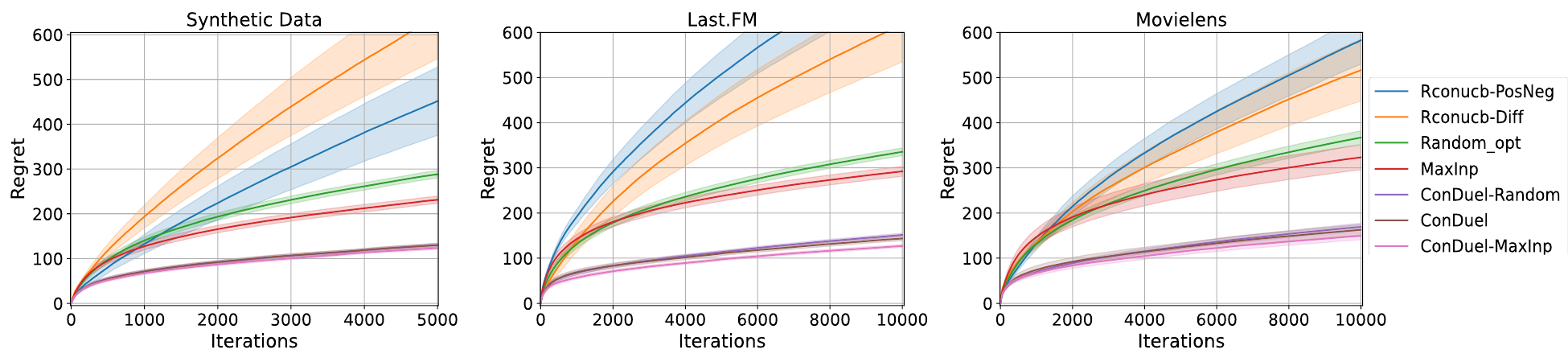}
    \caption{Cumulative regret on synthetic and real-world datasets}
    \Description{Show empirical evaluation for the algorithms}
    \label{fig:2}
\end{figure*}

\section{Experiments}
In this section, we describe experimental results on both synthetic data and real-world data to validate our proposed algorithm. The code is available at \url{https://github.com/shuashua0608/Con-Duel}. 
The arm-level and key-term level pairwise rewards are generated according to Eq. \eqref{eq:3} and Eq. \eqref{eq:4}, and the barycentric spanner $\mathcal{B}$ is computed in advance following \cite{awerbuch2004adaptive}. Specifically, we define the link function $\mu$ as the sigmoid function, thus leading to a logistic dueling bandit model. It should be noted that our algorithm can also be applied to other generalized linear model scenarios. 

\subsection{Implementation Details}
\noindent\textbf{Baselines}. We select the following algorithms as baselines to compare with ours:
\begin{itemize}
    \item \textbf{Random-opt}: A variant of MaxInp in \cite{saha2021optimal}, selecting two arms randomly from the constructed set $C_t$ without conversation from key-term level. This algorithm compares with MaxInp and shows the necessity of computing the "maximum informative pair" principle.
    \item \textbf{MaxInp} \cite{saha2021optimal}: A recently introduced algorithm designed for the contextual dueling bandits setting in GLM without a conversation mechanism. 
    \item \textbf{ConDuel-Random}: A variant of our algorithm that follows the same arm-pair selection principle but selects key-term pairs randomly.
    \item \textbf{ConDuel-MaxInp}: A variant of our algorithm that follows the same arm selection principle but selects key-term pairs with maximum information, that is, to choose $ k, k^{'} = \arg\max_{k, k^{'}\in\mathcal{K}}\|\tilde{x}_k-\tilde{x}_k^{'}\|_{M_t^{-1}}$ at $t$-th conversation.
\end{itemize}
Additionally, we compare our algorithms with RelativeConUCB from \cite{xie2021comparison}, namely, RelativeConUCB-Pos\&Neg and RelativeConUCB-Difference, which utilize relative feedback from key-term selection module in CRS. It should be noted that RelativeConUCB is designed for linear bandits and assumes that the arm-level model is a click model. For a fair comparison, we adapt their arm-level reward estimated from $r_a\sim Bernoulli(x_a^T\theta_*)$ to $r_a \sim Bernoulli(sigmoid(x_a^T\theta_*))$ for $a\in\mathcal{A}$ and update the regret as $R_T = \sum_{s=1}^T((x_{a_t^*}^T\theta_* - (x_{a_t}^T\theta_*))$ to fit in our problem setting.  We rewrite them as Rconucb-PosNeg and Rconucb-Diff, and give a general description of their key-term selection strategy:
\begin{itemize}
    \item \textbf{Rconucb-PosNeg}: Utilize relative feedback from the key-term level as two observations of absolute feedback: a positive observation of feedback 1 for the preferred key-term $\tilde{x_{k_1}}$ and a negative observation with feedback 0 for the less preferred key-term $\tilde{x}_{k_2}$;
    \item \textbf{Rconucb-Diff}: Incorporate the relative feedback as a single observation $(\tilde{x}_{k_1} - \tilde{x}_{k_2}, 1)$. 
\end{itemize}

\noindent\textbf{Metrics}. We use the cumulative dueling regret from Eq.\eqref{eq:5} to measure the performance of the algorithms, unless otherwise stated. Additionally, we plot the standard error for each algorithm to validate the stability of our proposed algorithm. We sequentially run the experiments ten times per user for each dataset and calculate the average cumulative regret for each algorithm.

\subsection{Synthetic Data}
\noindent\textbf{Data Generation.}  We construct the synthetic data following \cite{zhang2020conversational} and \cite{xie2021comparison}. 
First, we create a user set $\mathcal{U}$ with $|\mathcal{U}|$ = 200, a key-term set $\mathcal{K}$ with $|\mathcal{K}|$ = 500 and an arm set $\mathcal{A}$ with $|\mathcal{A}|$ = 5000, with the dimension of feature space to be $d=10$. We generate each element in user preference vector $\theta_u^*$ and arm feature vector $x_a$ independently from the standard normal distribution $N(0,1)$. Without loss of generosity, we normalize $\|\theta_u^{*}\|_2 = 1$ and $\|x_a\|_2 =1$.
To construct the weight matrix $W = [w_{a,k}]$, we follow similar procedures in \cite{wang2023efficient}: (1) We select an integer $n_k$ uniformly at random from [1, $M$], and select a subset of $n_k$ arms $\mathcal{A}_k$ to be related with key-term $k$. $M$ is set to be 10 in the experiments; (2) We assume each arm $a$ is related with $n_a$ key-terms subset $\mathcal{K}_a$ with equal weight $w_{a,k} = 1/n_a$, $\forall k\in \mathcal{K}_a $.  In the simulation, we set the time horizon $T$ = 5000, conversation frequency $b(t) = 10\lfloor\frac{t}{50}\rfloor$ and pool size $|A_t|$ = 50, unless otherwise stated. 
\begin{figure*}
    \centering   \includegraphics[width=.95\textwidth]{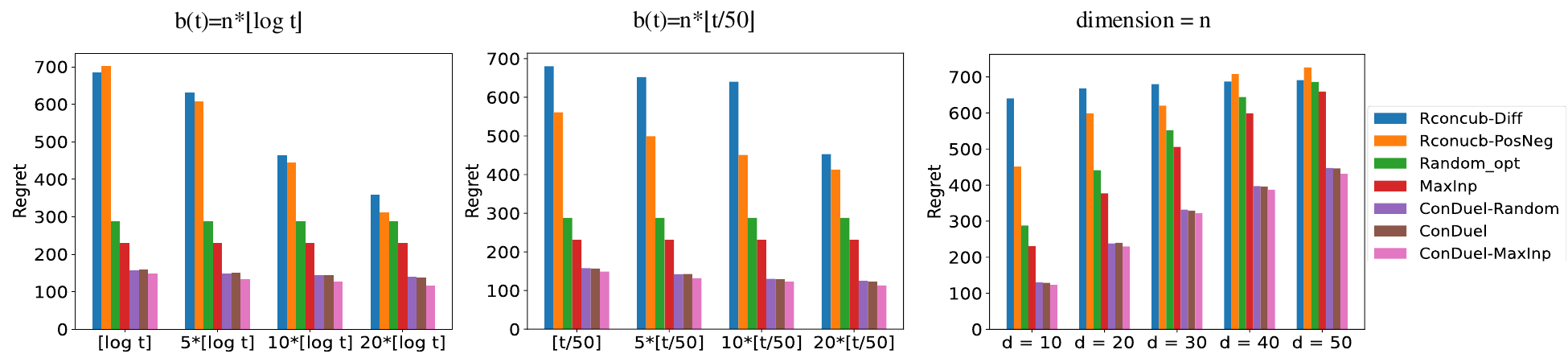}
    \caption{Ablation study on synthetic data}
    \Description{Empirical evaluation for our proposed algorithms}
 \label{fig:3}
\end{figure*}

\noindent \textbf{Experimental results.} The cumulative dueling regret curve with the standard error plot is shown in Figure~\ref{fig:2}. Our proposed algorithm ConDuel as well as its variants achieves better performances than MaxInp and Random-opt, realizing smaller regret and standard error. ConDuel and ConDuel-MaxInp perform slightly better than ConDuel-Random, indicating that carrying out explorative conversations can help reduce cumulative regret. Additionally, RelativeConUCB performs the worst compared with algorithms with nonlinear reward assumptions due to the less practical linear model assumption in the RelativeConUCB algorithm, which does not fit in with our experimental setting.

\noindent\textbf{Impact of conversation frequency and data dimension.}  We next study the impact of different conversation frequencies and data dimensions.  For the impact of conversation frequencies, since key-term-level conversations are less frequent than arm-level interactions, we consider the linear function: $b(t) = n \lfloor t/50\rfloor$, which means asking $n$ questions per 50 iterations, as well as the log function: $b(t) = n\lfloor \log t\rfloor$. We vary the value of $n$ to be 1, 5, 10, and 20 for both functions. The cumulative regrets of each conversation type and pool size are shown in Figure~\ref{fig:3}. According to Figure~\ref{fig:3}, more conversations can help reduce cumulative regrets more, for example, cumulative regret is the largest when $b(t) = \lfloor t/50\rfloor$ and smallest when $b(t) = 20\lfloor t/50\rfloor$. When the conversation frequency increases, our proposed algorithms utilizing explorative key-term strategy (namely, ConDuel and ConDuel-MaxInp) demonstrate more advantages than other algorithms.
To test the impact of data dimensions on our algorithms and validate the realizability of the proposed ConDuel in a higher dimension setting, we generate synthetic data of different dimensions, the data dimensions are set to be 20, 30, 40, and 50 with $b(t)=10\lfloor t/50\rfloor$. As is shown in Figure~\ref{fig:3}, as data dimension increases, the cumulative regret naturally increases, yet our proposed ConDuel algorithm with its variants still maintains superiority over other algorithms.

\subsection{Real-world Datasets}

\noindent\textbf{Data Generation.}  We next display the experimental results on two real-world datasets, Last.FM and Movielens, which are released in \cite{cantador2011second}. Last.FM is a dataset for music artist recommendations that contains 186,479 interaction records between 1,892 users and 17,632 artists. Movielens is a dataset that extends the original MovieLens10M dataset to include tagging information in IMDb and Rotten Tomatoes for movie recommendations and contains 47,957 interaction records between 2,113 users and 10,197 movies. 

We prepossess the data following \cite{xie2021comparison} and treat artists and movies as items, we then infer users’ real feedback on items based on the interaction records: if the user has assigned attributes to the item, the feedback is 1, otherwise, the feedback is missing. For both datasets, we extract $|\mathcal{A}|$ = 2,000 with the most assigned attributes by users and $N_u$ = 100 users who have assigned the most attributes. For each arm, we keep at most 20 attributes that are related to most items and treat them as the related key-terms of the item. All the kept key-terms associated with the arms form the key-term set $\mathcal{K}$—the number of key-terms for Last.FM is 2,726 and that for Movielens is 5,585. The weights of all key-terms related to the same arm are set to be equal, and we set the feature vectors to be $d=10$ to save the computation complexity. 

\noindent\textbf{Results.}  We compare the performance of algorithms in the two real datasets. We run the experiments 10 times and calculate all users' average regret over $T$ = 10,000 rounds on the fixed generated datasets. We set $b(t) = 10\lfloor\frac{t}{50}\rfloor$ and $|A_t|$ = 50. The evaluation results are shown in Figure~\ref{fig:2}. It can be seen from the figures that on both datasets, the ConDuel-MaxInp algorithm achieves the best performance in terms of cumulative regret and standard error, and the regret of ConDuel is also slightly lower compared with ConDuel-Random. In both cases, the ConDuel algorithms with explorative key-term selection strategy show their strengths compared to other algorithms. 

\section{Extension to MNL Bandit}

Besides pairwise comparison for conversational dueling bandits, we can also extend the conversational mechanism to the multiple comparison setting under the choice model, also known as the Multinomial Logit Bandit (MNL) problem~\cite{ou2018multinomial, oh2021multinomial}. The arm set $\mathcal{A}$ of size $N$ and key-term set $\mathcal{K}$ of size $K$ are defined in section~\ref{sec:prob}. We also define $\mathcal{C}$ to be the set of candidate assortments with size less than $q$, i.e. $\mathcal{C} = \{C\subset [N]:|C|\leq q\}$, where $q\geq 2$. In each iteration $t$, for the arm-level selection, the agent is offered an assortment ${C}_t = \{a_{i_1}, ..., a_{i_t}\}\subset\mathcal{C}$ and observes feature vector $x_{a, t}$ for each $a\in{C}_t$. The user purchase decision $o_t\in C_t\bigcup \{0\}$ is observed, and we can denote the user purchase decision for each $a\in{C}_t$ as $o_{a,t} = \textbf{1}(a\textit{ is chosen})\in\{0,1\}$ and $o_{0,t}$ indicating not choosing from the item set. Similarly, at $t$-th key-term level selection, the user observes key-term subset ${K}_t$ with $|K_t|\leq q$ and gives certain feedback $\Tilde{o}_t$, with $\Tilde{o}_{k,t}$ indicating whether key-term $k$ is chosen and $\Tilde{o}_{0,t}$ representing not choosing. 

\begin{algorithm}[!htp]
\SetAlgoLined
\KwIn{ $(\mathcal{A},\mathcal{K},\mathcal{W}),b(t), q$, initialization $T_0$;}
\textbf{Initialization}: 
\For {$t\in[T_0]$}{\eIf{$b(t)-b(t-1)>0$}{
$q_t=b(t)-b(t-1)$\;
\While{$q_t>0$}{
Randomly select $q$ key-terms from Barycentric Spanner $\mathcal{B}$\;
Update ${M}_t={M}_{t-1}+\sum_{k\in\mathcal{B}}\Tilde{x_{k,t}}\Tilde{x_{k,t}}^T$\;
$q_t= q_t - 1$;}}
{$M_t = M_{t-1}$}
Randomly choose $C_t\in\mathcal{C}$ with $|C_t| = q$\;
Update $M_t = M_{t-1} + \sum_{i\in{C}_t} x_{i,t} x_{i,t}^T$}
\For{$t = T_0+1,...,T$}{\eIf{$b(t)-b(t-1)>0$}{
$q_t=b(t)-b(t-1)$\;
\While{$q_t>0$}{
Offer ${K}_t$ based on key-term selection principle, and observe key-term level feedback $\Tilde{o}_t$\;
Update ${M}_t={M}_{t-1}+\sum_{k\in{K}_t}\Tilde{x_{k,t}}\Tilde{x_{k,t}}^T$\;
$q_t= q_t - 1$;}}
{$M_t = M_{t-1}$}
 MLE $\theta_t$ is estimated according to the regularized log-likelihood function in Eq~\ref{eq: 19}\;
Offer $C_t=\arg\max_{C\in\mathcal{C}} \hat{R}_t(C,\theta_t)$ to the user and observe user choice $o_t$\;
Update $M_t = M_{t-1} + \sum_{i\in{C}_t} x_{i,t} x_{i,t}^T$}
\caption{The ConMNL Algorithm}
\label{alg2}
\end{algorithm}
We define the ConMNL algorithm in \textbf{Algorithm}~\ref{alg2}. The user selection for item $a\in{C}_t$ and key-term $k\in{K}_t$ at round $t$ is given by the MNL choice model, defined in the following equations:
$$p_i(C_t,\theta_*) = \left\{
\begin{aligned}
    \frac{\exp(x_{i,t}^T\theta_*)}{1+ \sum_{j\in{C}_t}\exp(x_{j,t}^T\theta_*)}, \quad & i\in C_t,\\
     \frac{1}{1+ \sum_{j\in{C}_t}\exp(x_{j,t}^T\theta_*)}, \quad & i= 0,
\end{aligned}
\right.$$
$$ \Tilde{p}_t({K}_t,\theta_*)  = \left\{
\begin{aligned}
    \frac{\exp(\Tilde{x}_{k,t}^T\theta_*)}{1+ \sum_{j\in{K}_t}\exp(\Tilde{x}_{j,t}^T\theta_*)}, \quad & k\in K_t,\\
   \frac{1}{1+ \sum_{j\in{K}_t}\exp(\Tilde{x}_{j,t}^T\theta_*)},\quad & k= 0.
\end{aligned}
\right.$$
Here $\theta_*\in\mathbb{R^d}$. Notice that when $q=2$, the choice model can be seen as the dueling bandit model. We can rewrite the arm-level and key-term level choice model as follows:
\begin{align}\label{eq: 15}
    o_{a,t} = p_a(C_t,\theta_*) + \epsilon_{a,t}
\end{align}
\begin{align}\label{eq: 16}
    \Tilde{o}_{k,t} = \Tilde{p}_k(K_t,\theta_*) +\Tilde{\epsilon}_{k,t}
\end{align}
It is easy to verify that the noise $\epsilon_{a,t}$ and $\Tilde{\epsilon}_{k,t}$ are $\sigma^2$ sub-Gaussian variable with $\sigma = 0.5$. 
We assume that and the expected revenue of the assortment $C_t$ to be $R_t(C_t,\theta_*) = \sum_{j\in C_t}r_{j,t}p_j(C_t,\theta_*)$, where $r_{j,t}$ is the revenue from the recommendation if item $i$ is chosen by user at round $t$; and the optimal assortment $C_t^{*} = \arg\max_{C\in \mathcal{C}}R_t(C,\theta_*)$. The cumulative expected regret over time $T$ is defined as:
\begin{align}
    R(T) &= \sum_{t=1}^T\left(\sum_{j\in C_t^{*}}r_{j,t}p_j(C_t^{*},\theta_*)-\sum_{j\in C_t}r_{j,t}p_j(C_t,\theta_*)\right)\notag\\
    & = \sum_{t=1}^T \left(R_t(C_t^*,\theta_*) - R_t(C_t,\theta_*)\right).
\end{align}
Following the definition of Barycentric Spanner $\mathcal{B}$ of key-term set in sec \cref{sec:keyterm}, we assume $\lambda_{\mathcal{B}}^{'} = \lambda_{\min}\left(E_{k\sim \mathcal{B}}[\Tilde{x}_k\Tilde{x}_k^T]\right)>0$. Based on \cite{filippi2010parametric, li2017provably} and \cite{oh2021multinomial}, we also make the following assumptions:
\begin{figure*}
    \centering
    \vspace{-1mm}    \includegraphics[width=.95\textwidth]{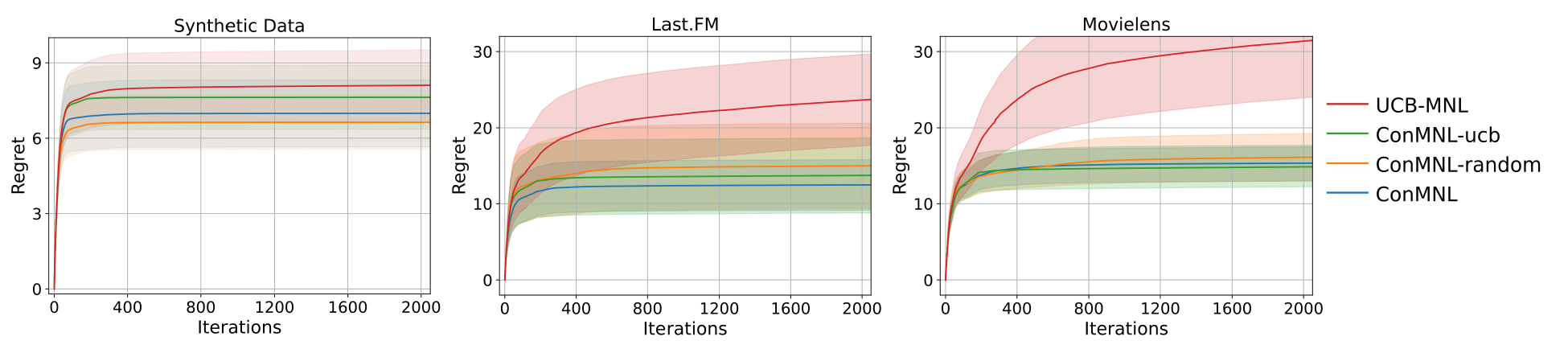}
    \caption{Cumulative regret on synthetic and real-world datasets}
    \Description{Empirical evaluation for the proposed algorithms}
    \label{fig:4}
\end{figure*}
\begin{assumption} \label{ass3}
For every $t$ and $i\in C$, there exists a constant $\kappa_2>0$, where $\kappa_2: = \min_{\|\theta-\theta_*\| \leq 1,|C|\leq q}p_i(C,\theta)p_0(C,\theta)$.
\end{assumption}
\begin{assumption}
Each feature vector $x_{i,t}$, $\|x_{i,t}\|\leq 1$ and there exists a constant $\sigma_0>0$, with $E[x_{i,t}^Tx_{i,t}]\geq \sigma_0$.
\end{assumption}

We estimate parameter $\theta_*$ following similar procedures in sec~\cref{sec:pipe}. The log-likelihood function till $t$-th round under parameter $\theta$ is given by
\begin{align}\label{eq: 18}
    L_t(\theta) = \sum_{\tau=1}^{t-1}\sum_{i\in {C}_{\tau}}o_{s,i}log(\mu_i({C}_{\tau}, \theta))+\sum_{\tau=1}^t\sum_{k\in{K}_\tau}\Tilde{o}_{\tau,k}log(\Tilde{\mu}_k({K}_\tau, \theta).
\end{align}
Setting $\nabla_\theta L_t(\theta)=0$, the maximum likelihood estimation $\theta_t$ is the solution of:
\begin{align}\label{eq: 19}
    \sum_{\tau=1}^{t-1}\sum_{i\in {C}_{\tau}}(o_{i,s}-\mu_i({C}_{\tau}, \theta))x_{i,s}+ \sum_{\tau=1}^t\sum_{k\in{K}_\tau}(\Tilde{o}_{k,\tau}-\Tilde{\mu}_k({K}_\tau, \theta))\Tilde{x}_{k,\tau} = 0.
\end{align}
Define $M_t = \sum_{\tau=1}^{t-1}\sum_{i\in {C}_{\tau}}x_{i,\tau}x_{i,\tau}^T+ \sum_{\tau=1}^t\sum_{k\in{K}_\tau}\Tilde{x}_{k,\tau}\Tilde{x}_{k,\tau}^T $.
We can calculate the MLE $\theta_t$ to obtain the UCB estimate $z_{a,t}=x_{a,t}^T\theta_t + \alpha_t\|x_{a,t}\|_{M_t^{-1}}$ regarding the utility of each $a\in\mathcal{A}_t$ at time $t$, with $\alpha_t=\frac{1}{2\kappa_2}\sqrt{2d\log(1+\frac{b(t) + t}{d})+\log(\frac{1}{\delta})}$. In the item selection module, we construct the optimal estimate of the expected revenue by choosing $C$ as 
\begin{align}
    \hat{R}_t(C) = \frac{\sum_{i\in C}u(i,t)\exp(z_{i,t})}{1 +\sum_{i\in C}\exp(z_{i,t})},
\end{align}
and offer $C_t =\arg\max_{C\in\mathcal{C}} \hat{R}_t(C)$ to the user at time $t$. For the key-term query module, we choose $q$ key-terms with each uniformly sampled from barycentric spanner $\mathcal{B}$ to form $\mathcal{K}_t$. 

The regret upper bound of ConMNL is given in Theorem~\ref{th2}.
\begin{theorem}\label{th2}
Assume conversation frequency to be $b(t) = b \cdot t$, with $b\in(0,1)$, and $\alpha_t = \frac{1}{2\kappa_2}\sqrt{2d\log(1+\frac{b(t) + t}{d})+ 2\log(t)}$, and $T_0 = \mathcal{O}\left(max\{\frac{1}{\kappa_2^2}(d\log(\frac{b(T)+T}{d}+4\log(T)), q/\sigma^2, \frac{256}{bq\lambda_{\mathcal{B}}^{'}}\log(\frac{128d}{\delta\lambda^2_{\mathcal{B}^{'}}}) \}\right)$. The expected regret of ConMNL is upper bound by 
$R_T = \mathcal{O}(\sqrt{dTq\log(T)}).$
\end{theorem}

\subsection{Experiments}
We compare our ConMNL algorithm with the following baselines on the previous three datasets:  
\begin{itemize}
    \item \textbf{UCB-MNL}. An algorithm designed for MNL contextual bandit in \cite{oh2021multinomial} with no conversation on key-terms.
    \item \textbf{ConMNL-ucb}: A variant of ConMNL that selects $q$ key-terms at each conversation based on UCB estimate $\Tilde{x}_{k,t}^T\theta_t + \alpha_t\|\Tilde{x}_{k,t}\|_{M_t^{-1}}$.
    \item \textbf{ConMNL-random}: A variant of ConMNL that selects $q$ key-terms randomly at each conversation.
\end{itemize}
We set horizon $T = 2000$ and $q=4$ for each dataset, allowing at most $q$ items and key-terms for the user to choose from. We also assume the expected revenue of choosing item $i$ is given by $r_{i,t} = x_{i,t}^T\theta_*$, the utility of each item. The conversation frequency $b(t)$ is set to be $5\lfloor\frac{t}{50}\rfloor$, and pool size $|A_t|$ = 50. It should be noted that the ConMNL-ucb and ConMNL-random follow the same item assortment selection principle proposed in the ConMNL algorithm, aiming to compare the impact of different conversation mechanisms on the performance of recommender systems. We ran experiments on each user 10 times and calculated the average regret as well as standard error. The regret curve for each dataset is shown in Fig~\ref{fig:4}.\\
It can be seen from the figures that our algorithm along with its variants all perform better than UCB-MNL on each dataset, confirming that carrying out conversations on key-terms can enhance the model performance. Furthermore, ConMNL performs the best on Last.FM dataset, while ConMNL-ucb achieves relatively better results on both synthetic dataset and Movielens dataset. This may be due to the complexity of constructing a $q$-size choice set, where utilizing information solely from the barycentric spanner subset from the key-term set may be inadequate to capture the user preference. 

\section{Conclusion}
In this paper, we study a novel framework of conversational bandits with an informative feedback mechanism in the generalized linear models and propose the ConDuel algorithm that can guarantee relative feedback from both key-term and item modules. We design new methods to effectively duel key-term pairs and item pairs in our algorithm, which allow the system to conduct exploratory conver- \vfill\eject\noindent sations to utilize key-term pairwise feedback in the key-term module. Meanwhile, we select the most informative pairs in the item module to grasp the user preferences more accurately. We prove a regret upper bound of $\mathcal{O}(\sqrt{d T\log(T)})$ of our algorithm, and extensive experiments on both synthetic data and real-world datasets have demonstrated the competitiveness of our algorithm. We also extend our algorithm to multiple comparisons under the MNL choice model and propose the ConMNL algorithm with a theoretical guarantee. For future research, it would be intriguing to consider: 1) Incorporate additional structure for key-term module, such as knowledge graph \cite{zhao2022knowledge} or clustering \cite{wu2021clustering}; 2) Consider different distributions on key-terms and higher dimensions of large-scale dataset in real dataset experiments.

\section*{Acknowledgments}
Mengdi Wang acknowledges the support by NSF grants DMS-1953686, IIS-2107304, CMMI-1653435, ONR grant 1006977, and C3.AI.

\clearpage
\bibliographystyle{ACM-Reference-Format}
\balance
\bibliography{sample-base}
\appendix

\section{Proofs for ConDuel Algorithm}

\begin{proof}[Proof of Lemma 1]

According to Eq.~\eqref{eq:7} and Eq.~\eqref{eq:8} and the definition of $g_t(\theta)$, $\forall \theta_1,\theta_2\in \Theta$, we have:
{\small\begin{align}
g_t(\theta_1)-g_t(\theta_2) & =(G_t^a(\theta_1,\theta_2)+G_t^k(\theta_1,\theta_2)+\lambda I)(\theta_1-\theta_2)\notag\\
&\triangleq G_t(\theta_1,\theta_2)(\theta_1-\theta_2)>\kappa_1 M_t(\theta_1-\theta_2),
\end{align}}
where $G_t^a(\theta_1,\theta_2) = \sum_{s=1}^{t-1} \mu^{'}(d_s^T\Bar{\theta}(\theta_1,\theta_2))d_s d_s^T$,
and $G_t^k(\theta_1,\theta_2) =\sum_{s=1}^t\sum_{\Tilde{d}\in\Tilde{\mathcal{D}}_s}\mu^{'}(\Tilde{d}_{s}^T\Bar{\theta}(\theta_1,\theta_2)\Tilde{d}_{s}\Tilde{d}_{s}^T$.

Based on the mean value theorem, we also have $\Bar{\theta}(\theta_1,\theta_2) = v\theta_1+(1-v)\theta_2$.  According to Eq.~\eqref{eq:3} and Eq.~\eqref{eq:4}, we have the equality:\vspace{-1em}
{\small\begin{align}
    g_t(\theta_t) - g_t(\theta_*) & = \sum_{s=1}^{t-1} \epsilon_s d_s 
+\sum_{s=1}^t\sum_{\Tilde{d}\in\Tilde{\mathcal{D}}_s}\Tilde{\epsilon}_s\Tilde{d}_{s}-\lambda\theta_*\notag\\
    & \triangleq S_t-\lambda\theta_*>\kappa_1 M_t(\theta_t-\theta_*)
\end{align}}

Combined with Eq.~\eqref{eq:7}, \eqref{eq:8} and \eqref{eq:9}, $\forall x\in \mathbb{R}^d$, we thus have:
{\small\begin{align}
   &|x^T(\theta_t^{(1)}-\theta_*)|= |x^T G_t^{-1}(\ \theta_t^{(1)},\theta_*)(g_t(\theta_t^{(1)})-g_t(\theta_*))|\notag\\
   &\leq \|x\|_{G_t^{-1}(\theta_t^{(1)},\theta_*)}\|g_t(\theta_t^{(1)})-g_t(\theta_*)\|_{G_t^{-1}(\theta_t^{(1)},\theta_*)}\notag\\
    &\leq^{(1)} \frac{1}{\kappa_1}\|x\|_{M_t^{-1}}\|g_t(\theta_t^{(1)})-g_t(\theta_*)\|_{M_t^{-1}}\notag\\
    &= \frac{1}{\kappa_1}\|x\|_{M_t^{-1}}\|g_t(\theta_t^{(1)})-g_t(\theta_t)+g_t(\theta_t)-g_t(\theta_*)\|_{M_t^{-1}}\notag\\
    &\leq^{(2)} \frac{1}{\kappa_1}\|x\|_{M_t^{-1}}(\|g_t(\theta_t^{(1)})-g_t(\theta_t)\|_{M_t^{-1}}+\|g_t(\theta_t)-g_t(\theta_*)\|_{M_t^{-1}})\notag\\
    &\leq^{(3)} \frac{2}{\kappa_1}\|x\|_{M_t^{-1}}\|g_t(\theta_t)-g_t(\theta_*)\|_{M_t^{-1}}\notag\\
    &\leq \frac{2}{\kappa_1}(\|S_t\|_{M_t^{-1}}+\sqrt{\lambda\kappa_1}\|\theta_*\|_2)\|x\|_{M_t^{-1}}
\end{align}}
The first inequality $(1)$ comes from $G_t(\theta_t^{(1)},\theta_*)\geq \kappa_1 M_t>0$, and therefore $G_t(\theta_t^{(1)},\theta_*)^{-1}\leq \frac{1}{\kappa_1}M_t$. The second inequality is the application of triangle inequality, and the second inequality $(3)$ is based on the definition of $\theta_t^{(1)}$ from Eq.~\eqref{eq:9}.

Notice that $\epsilon_t$ and $\Tilde{\epsilon}_t$ are $R$-subgaussian, and $\forall t$, $\|d_t\|_2, \|\Tilde{d}_t\|_2\leq 2$. According to Theorem 1 in \cite{abbasi2011improved}, we have:
{\small\begin{align}
    \|S_t\|_{M_t^{-1}}&\leq R\sqrt{2\log(\frac{\det(M_t)^{1/2}/(\frac{\lambda}{\kappa_1})^{d/2}}{\delta})}\notag\\
    &\leq R\sqrt{d\log((1+\frac{4\kappa_1 (t+b(t))}{\lambda d})/\delta)}.
\end{align}}
Therefore, $\alpha_t \triangleq \frac{2}{\kappa_1}(R\sqrt{d\log((1+\frac{4\kappa_1(t+ b(t))}{d\lambda})/\sigma)}+\sqrt{\lambda\kappa_1}\|\theta_*\|_2).$
\end{proof}

\begin{proof}[Proof of Lemma 2]

Lemma 2 follows the existing results of Proposition 1 in \cite{li2017provably}.

Denote $\mathcal{B}$ as the barycentric spanner for key-term set $\mathcal{K}$. Let $x,y$ be random vectors sampled independently and uniformly from $\mathcal{B}$, i.e. $x, y\overset{iid}{\sim} \mathcal{B}$. Define $\Sigma \triangleq E_{x,y \overset{iid}{\sim} \mathcal{B}}[(x-y) (x-y)^T]$. For ease of understanding, we assume the pair of key-terms at round $t$ conversation with contextual vectors $(\Tilde{x}_t,\Tilde{y}_t)$, and the key-term level design matrix is denoted as $\Tilde{M}_t = \sum_{s=1}^t\sum_{\Tilde{x},\Tilde{y}\overset{iid}{\sim} \mathcal{B}}(\Tilde{x}_s-\Tilde{y}_s)(\Tilde{x}_s-\Tilde{y}_s)^T$.

Define $z_t= \Sigma^{-1/2}(x_t-y_t)$, then $z_t$ is isotropic, namely, $E[z_t z_t^T] = I$. Define $U_t = \sum_{s=1}^{t}\sum_{\Tilde{x}_s,\Tilde{y}_s\overset{iid}{\sim} \mathcal{B}} z_s z_s^T = \Sigma^{-1/2}\Tilde{M}_t\Sigma^{-1/2}$. 

From \textbf{Lemma 1} in \cite{li2017provably}, with probability at least ($1-2\exp(-C_2 x^2)$):
$$\lambda_{\min} (U_t)\geq b(t) - C_1\sigma^2\sqrt{b(t) d}-\sigma^2 x\sqrt{b(t)},$$
where $\sigma$ is the sub-gaussian parameter of $z$ and is upper-bounded by $\|\Sigma^{-1/2}\| = \lambda_{\min}(\Sigma)^{-1/2}$. Therefore, with probability at least $(1-\delta)$: 
$$\lambda_{\min} (U_t)\geq b(t) - \frac{1}{\lambda_{\min}(\Sigma)}(C_1\sqrt{b(t) d}+ x\sqrt{b(t)}).$$
Furthermore, the minimum eigenvalue of $\Tilde{M}_t$ is bounded as follows:
\begin{align}
    &\lambda_{min}(\Tilde{M}_t) = \min_{x\in\mathbb{B}^d}x^T \Tilde{M}_t x = \min_{x\in\mathbb{B}^d}x^T \Sigma^{1/2}U_t  \Sigma^{1/2} x \geq \lambda_{\min}(U_t)\lambda_{\min}(\Sigma)\notag\\
    &\geq  \lambda_{\min}(\Sigma)(b(t) - \lambda_{\min}(\Sigma)^{-1}(C_1\sqrt{b(t) d}+C_2\sqrt{b(t)\log(1/\delta)}\notag\\
    & = \lambda_{\min}(\Sigma)b(t) -(C_1\sqrt{d}+C_2\sqrt{\log(1/\delta)})\sqrt{b(t)}.
\end{align}
We denote $\lambda_{\min}(\Sigma) = \lambda_B$. When $b(t)\geq \frac{4(C_1\sqrt{d}+C_2\sqrt{\log(1/\delta))^2}}{\lambda_B^2}$, we have:\vspace{-1em}
\begin{align}
    \lambda_{\min}(\Tilde{M}_t)\geq \frac{\lambda_B b(t)}{2}.
\end{align}
When the conversation frequency function is linear, i.e., $b(t) = b\cdot t$ for some $b\in(0,1)$, then with probability at least $(1-\delta)$, we have: $$\lambda_{\min}(\Tilde{M}_t)\geq \frac{\lambda_B bt}{2},$$ 
as long as $t\geq  \frac{4(C_1\sqrt{d}+C_2\sqrt{\log(1/\delta))^2}}{b\lambda_B^2} \triangleq t_0$.
\end{proof}

\begin{proof}[Proof of Lemma 3]

According to Lemma~\ref{lemma:2},
{\small$$\lambda_{\min}(M_t)\geq \lambda_{\min}(\Tilde{M}_t)+ \frac{\lambda}{\kappa_1}\geq \frac{\lambda_B b( t)}{2}+\frac{\lambda}{\kappa_1}$$ when $b(t)\geq \frac{4(C_1\sqrt{d}+C_2\sqrt{\log(1/\delta))^2}}{\lambda_B^2}$}.

We have assumed that $\|x_t\|_2\leq1$, $\forall x_t\in\mathcal{A}_t$, and 
$\|d_t\|_2 = \|x_t-x_t^{'}\|_2 \leq \|x_t\|_2 +\|x_t^{'}\|_2 \leq 2$, where $(x_t, x_t^{'})$ denote the contextual vectors for the selected pair of arms at round $t$. Therefore we can obtain the following inequality: 

$\|d_t\|_{M_t^{-1}}\leq \sqrt{\frac{1}{\lambda_{\min}(M_t)}}\|d_t\|_2\leq 2\sqrt{\frac{1}{\lambda_{\min}(M_t)}}\leq 2({\frac{\lambda_B b (t)}{2}+\frac{\lambda}{\kappa_1}})^{-1/2}$, as well as:
{\small\begin{align}\label{eq:14}
    &\sum_{s = t_0+1}^t\|d_s\|_{M_s^{-1}}\leq 2\sum_{s = t_0+1}^t\sqrt{\frac{1}{\lambda_{\min}(M_s)}}\leq 2\sum_{s = t_0+1}^t(\frac{\lambda_B b (s)}{2}+\frac{\lambda}{\kappa_1})^{-1/2}\notag\\
    &\leq 2\sum_{s = t_0+1}^t(\frac{\lambda_B b (s)}{2})^{-1/2}\leq 2 \int_{s=t_0}^t (\frac{\lambda_B b (s)}{2})^{-1/2}d s.
\end{align}}

Though the upper bound given in \eqref{eq:14} is complicated, when the conversation frequency function is linear, i.e., $b(t) = b \cdot t$ for some random $b\in (0,1)$, we can easily calculate the following inequality:
{\small\begin{align}
    &\sum_{s = t_0+1}^t\|d_s\|_{M_s^{-1}}\leq  8 \sqrt{\frac{s}{2b\lambda_B}}\bigg |_{s=t_0}^{t}= 8(\sqrt{\frac{t}{2b\lambda_B}}-\sqrt{\frac{t_0}{2b\lambda_B}})\leq 8\sqrt{\frac{t}{2b\lambda_B}}
\end{align}}
\end{proof}

\begin{proof}[Proof of Theorem 1]

This proof lies in expressing the regret bound in terms of the above concentration results from Lemma ~\ref{lemma:1}, and it is possible owing to the arm selection strategy which follows the \textit{most informative pair} strategy from \cite{saha2021optimal}. Suppose we have selected a pair of arms at round $t$ with contextual vector being $(x_t,x_t^{'})$, and assume $x_{t}^{*} = \arg\max_{a\in\mathcal{A}_t} x_{a,t}^T\theta_{*}$, then we have:
{\small\begin{align}
2r_t=&({x_t^{*}}^T\theta_{*}-x_{t}^{T}\theta_{*})+({x_t^{*}}^{T}\theta_{*}-{x_{t}^{'}}^T\theta_{*})\notag\\
&\leq \bigg((x_{t}^{*}-x_t)^T \theta_{*}+(x_{t}^{*}-x_t^{'})^T\theta_* \bigg)\notag\\
& = \bigg( (x_{t}^{*}-x_t)^T(\theta_*-\theta_t^{(1)})+(x_{t}^{*} - x_t)^T \theta_t^{(1)}\notag\\
&+ (x_{t}^{*}-x_t^{'})^T(\theta_*-\theta_t^{(1)})+(x_{t}^{*} - x_t^{'})^T \theta_t\bigg)\notag\\
&\leq^{(1)} \bigg(\alpha_t\|x_{t}^{*}-x_t\|_{M_t^{-1}}+\|\theta_*-\theta_t^{(1)}\|_{M_t}\|x_{t}^{*}-x_t\|_{M_t^{-1}}\notag\\
&+\|\theta_*-\theta_t^{(1)}\|_{M_t}\|x_{t}^{*}-x_t^{'}\|_{M_t^{-1}}+ \alpha_t\|x_{t}^{*}-x_t^{'}\|_{M_t^{-1}}\bigg)\notag\\
&\leq^{(2)}\bigg( 2\alpha_t\|x_{t}^{*}-x_t\|_{M_t^{-1}} + 2\alpha_t\|x_{t}^{*}-x_t^{'}\|_{M_t^{-1}}\bigg)\notag \leq^{(3)} 4\alpha_t\|x_t-x_t^{'}\|_{M_t^{-1}},
\end{align}}

where the first inequality (1) holds due to the construction of $\mathcal{C}_t$ and the fact that both $x_t,x_t^{'}\in\mathcal{C}_t$, so that $\left|(x_t^* - x_t)^T \theta_t^{(1)}\right| \leq \alpha_t \|x_t^* - x_t\|_{M_t^{-1}}$ and $\left|(x_t^* - x_t^{'})^T \theta_t^{(1)}\right| \leq \alpha_t \|x_t^* - x_t^{'}\|_{M_t^{-1}}$. Inequality (2) follows from Lemma 1, where we have proved $\alpha_t$ is the upper bound for $\|\theta_*-\theta_t^{(1)}\|_{M_t^{-1}}$. The last inequality comes from the arm selection strategy.

Denote $d_t = x_t-x_t^{'}$, combined with the definition of $\alpha_t$, we have
{\small$$r_t\leq \frac{4}{\kappa_1}(R\sqrt{d\log((1+\frac{4\kappa_1(t+ b(t))}{d\lambda})/\sigma)}+\sqrt{\lambda\kappa_1}\|\theta_*\|_2)\|d_t\|_{M_t^{-1}}.$$}

Therefore, the cumulative regret over time $T$ is:
{\small\begin{align}
    R(T) & =\sum_{t=1}^{t_0}r_t+\sum_{t=t_0+1}^T r_t\notag\leq t_0+\sum_{t=t_0+1}^T r_t
    \\
    &\leq t_0+2 \alpha_T\sum_{t=t_0+1}^{T}\|d_t\|_{M_t^{-1}}\leq t_0+ 2 \alpha_T\sum_{s = t_0+1}^t(\frac{\lambda_B b (s)}{2})^{-1/2}\notag\\
    &\leq t_0 +  2\alpha_T\int_{s=t_0}^t (\frac{\lambda_B b (s)}{2})^{-1/2}d s.
\end{align}}

When $b(t) = b\cdot t$, we have:
\begin{align}
    R(T) &\leq t_0+\frac{32}{\kappa_1}\bigg(R\sqrt{d\log((1+\frac{4\kappa_1(T+ b T)}{d\lambda})/\sigma)}\notag\\
    &+\sqrt{\lambda\kappa_1}\|\theta_{*}\|_2\bigg)\sqrt{\frac{T}{2b\lambda_B}}= \mathcal{O}(\frac{1}{\kappa_1}\sqrt{d T\log(T)}).
\end{align}

\end{proof}

\section{Intuition for ConDuel-MaxInp Algorithm}
We design the ConDuel-MaxInp algorithm based on intuition as follows:

Denote $n_t = \lfloor b(t)- b(t-1) \rfloor$ as the number of conversations between the agent and the user when $q(t) = 1$, and $n_t = 0$ when $q(t)=0$ for the key-term selection module. Based on Eq~\eqref{eq:8}, we rewrite $M_t$ as:
\begin{equation}
    M_t = \sum_{s=1}^t d_t d_t^T +\sum_{s=1}^t\sum_{j=1}^{n_s} \Tilde{d}_j\Tilde{d}_j^T + \lambda/\kappa_1 I,
\end{equation}
and also define $M_{t,j} = M_{t-1} + d_{t-1}d_{t-1}^T + \sum_{i=1}^{j-1}\Tilde{d}_i\Tilde{d}_i^T$ for $j \in \{1,...,n_t\}$ when $n_t\neq 0$, then it is easy to obtain the following equation:
\begin{align}
    det(M_t) & = det(M_{t-1})(1+\|d_{t-1}\|^2_{M_{t-1}^{-1}})\prod_{j=1}^{n_t}(1+\|\Tilde{d}_j\|^2_{M_{t,j}^{-1}})\notag\\
    & = det(M_0)\prod_{s=1}^{t-1}(1+\|d_{s}\|^2_{M_{s}^{-1}})\prod_{s=1}^{t}\prod_{j=1}^{n_s}(1+\|\Tilde{d}_j\|^2_{M_{s,j}^{-1}}).\notag
\end{align}
Notice that $\frac{1}{2}x\leq log(1+x)\leq x $ for $x\in[0,1]$, we have:

\begin{align}
    &\sum_{s=1}^{t-1}\|d_s\|_{M_s^{-1}}^2+\sum_{s=1}^t\sum_{j=1}^{n_s}\|\Tilde{d}_{j}\|^2_{M_s^{-1}}\notag\\
    &\leq \sum_{s=1}^{t-1}\|d_s\|_{M_s^{-1}}^2+\sum_{s=1}^t\sum_{j=1}^{n_s}\|\Tilde{d}_{j}\|^2_{M_{s,j}^{-1}}\leq 2\log\bigg(\frac{det(M_t)}{det(M_0)}\bigg).
\end{align} 

Therefore, when applying the ''Maxinp'' strategy on selecting key-terms in the ConDuel-Maxinp algorithm, i.e., choosing key-term satisfying $\Tilde{k}_{t} = \arg\max_{\Tilde{d}\in\Tilde{D}_t}\|\Tilde{d}\|^2_{M_t^{-1}}$ at $t$ conversation, the system carries out explorative conversations and reduces uncertainty on key-terms.

\section{Proofs for ConMNL Algorithm}
We start by giving the following lemmas to prove Theorem 2.

\begin{lemma}\label{lemma:4}(\cite{kveton2020randomized}, Lemma 9) If  $\lambda_{\min}(M_{T_0})\geq max\{\sigma^2\kappa_2^{-2}(d\log(\frac{b(T)+T}{d})+2\log(\frac{1}{\delta}),q\}$, then $\forall t\geq T_0$, with probability at least $(1-\delta)$, we have $$\|\theta_t - \theta_*\|\leq 1.$$
\end{lemma}

\begin{lemma}\label{lemma:5}
Suppose $\|\theta_t-\theta_*\|\leq 1$ for $t\geq T_0$, then with probability at least $(1-\delta)$, we have
\begin{align}
    \|\theta_t-\theta_*\|_{M_t}\leq \frac{1}{2\kappa_2}\sqrt{2d\log(1+\frac{b(t) + t}{d})+\log(\frac{1}{\delta})}\triangleq\alpha_t.
\end{align}
\end{lemma}

\begin{lemma}\label{lemma:6}
Following the definition of Barycentric Spanner $\mathcal{B}$ of key-term set, we assume $\lambda_{\mathcal{B}}^{'} = \lambda_{\min}\left(E_{k\sim \mathcal{B}}[\Tilde{x}_k\Tilde{x}_k^T]\right)>0$. Define $b(t) = b t$ for $b\in (0,1)$, then $\forall t\geq \frac{256}{bq\lambda_{\mathcal{B}}^{'}}\log(\frac{128d}{\delta\lambda^2_{\mathcal{B}^{'}}})\triangleq t_0$, and $\delta\in(0,1/8]$, with probability at least $(1-\delta)$,  we have
$$\sum_{s = t_0+1}^t \sum_{i\in C_s}\|x_{i,t}\|_{M_s^{-1}}\leq 2\sqrt{\frac{2qt}{b\lambda_{\mathcal{B}^{'}}}}$$
\end{lemma}

\begin{lemma}(\cite{oh2021multinomial}, lemma 4) \label{lemma:7}
With $C_t^{*}$ defined as the optimal assortment, and $C_t = \arg\max_{C\in\mathcal{C}}\hat{R}_t(C,t)$. If $z_{i,t}\geq x_{i,t}^T\theta_*$ for every $i\in C_t^{*}$, then we have
$$R_t(C_t^{*},\theta_*)\leq \hat{R}_t(C_t^{*})\leq \hat{R}_t(C_t).$$
\end{lemma}

\begin{lemma}\label{lemma:8}(\cite{oh2021multinomial}, Lemma 3)
With $\alpha_t$ defined in Lemma~\ref{lemma:5}, suppose $z_{a,t} = x_{a,t}^T\theta_t + \alpha_t\|x_{a,t}\|_{M_t^{-1}}$ for all $a\in\mathcal{A}_t$, then we have
$$0\leq z_{a,t} - x_{a,t}^T\theta_* \leq 2\alpha_t\|x_{a,t}\|_{M_t^{-1}}.$$
\end{lemma}

\begin{proof}[Proof of Theorem 2]
When $T_0 = \mathcal{O}\left(max\{t_0, t_1, q/\sigma^2\}\right)$, with $t_0=\frac{256}{bq\lambda_{\mathcal{B}}^{'}}\log(\frac{128d}{\delta\lambda^2_{\mathcal{B}^{'}}})$, and $\frac{1}{\kappa_2^2}(d\log(\frac{b(T)+T}{d}+4\log(T))$, we have $\|\theta_t-\theta_*\|\leq 1$ by Lemma~\ref{lemma:4}, and the regret becomes:
\begin{align}
    R_T&=\sum_{t=1}^{T_0} \left(R_t(C_t^*,\theta_*) - R_t(C_t,\theta_*)\right) + \sum_{t=T_0+1}^T  \left(R_t(C_t^*,\theta_*) - R_t(C_t,\theta_*)\right)\notag\\
    &\leq T_0 +  \sum_{t=T_0+1}^T\left(\hat{R}_t(C_t) - R_t(C_t,\theta_*)\right)+\sum_{t=1}^T \mathcal{O}(t^{-2})\notag\\
    &\leq T_0 + 2\alpha_T \sum_{t = T_0+1}^T\sum_{i\in C_t}\|x_{i,t}\|_{M_t^{-1}}+\mathcal{O}(1)\notag\\
    &\leq T_0 + 4\alpha_T\sqrt{\frac{2qt}{b\lambda_{\mathcal{B}^{'}}}}+\mathcal{O}(1)
\end{align}
Combined with Lemma~\ref{lemma:6} and the definition of $\alpha_T$, Theorem 2 is proved.
\end{proof}

\end{document}